%% file: main.tex
\documentclass{article} 
\usepackage{iclr2027_conference,times}

\input{math_commands.tex}

\usepackage{graphicx}
\usepackage{amsmath}
\usepackage{amssymb}
\usepackage{hyperref}
\usepackage{url}
\usepackage{booktabs}
\usepackage{tabularx}
\usepackage{svg}
\usepackage{float}

\title{From Chain-of-Thought to Loops: Non-Autoregressive Latent Reasoning via Looped Transformers}

\author{
    \parbox{\textwidth}{\centering
        \vspace*{0.15in} 
        Gerard Grau García\textsuperscript{1, 2} \qquad
        Arnau Padrés Masdemont\textsuperscript{1} \qquad
        Niccolò Grillo\textsuperscript{1, 3} \\[0.1cm]
        Jordi Ros Giralt\textsuperscript{1} \qquad
        Arash Behboodi\textsuperscript{1} \qquad
        Victor Conchello Vendrell\textsuperscript{1, 4}\thanks{Correspondence to: \texttt{vconchellovendrell@g.harvard.edu}, \texttt{apadres@qti.qualcomm.com} and \texttt{gerardg@mit.edu}. Work done while at Qualcomm AI Research.} \\[0.25cm]
        {\normalfont \textsuperscript{1}Qualcomm AI Research \qquad \textsuperscript{2}MIT \qquad \textsuperscript{3}University of Cambridge \qquad \textsuperscript{4}Harvard University}
    }
}

\usepackage{xcolor}
\definecolor{deltaGreen}{RGB}{20,150,50}
\definecolor{deltaRed}{RGB}{190,55,55}
\definecolor{deltaGray}{RGB}{105,105,105}
\newcommand{\resgood}[1]{{\scriptsize\textcolor{deltaGreen}{(#1)}}}
\newcommand{\resbad}[1]{{\scriptsize\textcolor{deltaRed}{(#1)}}}
\newcommand{\resneutral}[1]{{\scriptsize\textcolor{deltaGray}{(#1)}}}

\usepackage{natbib}

\iclrfinalcopy 

\begin{document}

\maketitle

\begin{abstract}
Chain-of-thought (CoT) reasoning often improves language-model performance by
giving models additional computation before answering. However, explicit CoT
expresses this computation as a sequence of autoregressively generated tokens.
Latent reasoning replaces these tokens with compact continuous states, but most
autoregressive latent-reasoning methods retain a left-to-right dependency among
latent vectors. We introduce LLoCoT: a looped latent-reasoning framework that replaces
left-to-right latent generation with iterative refinement of a compact latent
workspace. A shared transformer is reapplied for a small number of refinement
iterations, jointly updating the latent slots based on the prompt and the
evolving workspace state. Using the refined state, a probabilistic head
predicts a distribution from which latent tokens are sampled in parallel and
used to condition an autoregressive decoder for answer generation. Training
uses continuous representations derived from explicit CoT together with a
final-answer prediction loss and likelihood-based supervision of the latent
states. Across HumanEval and MBPP, LLoCoT achieves the highest mean among the evaluated methods, performing on par in accuracy
with Reasoning SFT, our explicit-CoT baseline, while outperforming the base
model, answer-only SFT and NF-CoT. Relative to Reasoning SFT, LLoCoT reduces
time to the first answer token by approximately $36\times$ and
reasoning-phase latency by approximately $42\times$, while increasing
end-to-end throughput by $9.2\%$. This design replaces serial thought
generation with parallel latent-slot refinement while retaining probabilistic
latent modeling and autoregressive answer decoding.
\end{abstract}

\input{sections/introduction}


\vspace{-15pt}
\begin{figure}[htbp]
  \centering
  \makebox[\linewidth][c]{%
    \includegraphics[width=1\linewidth]{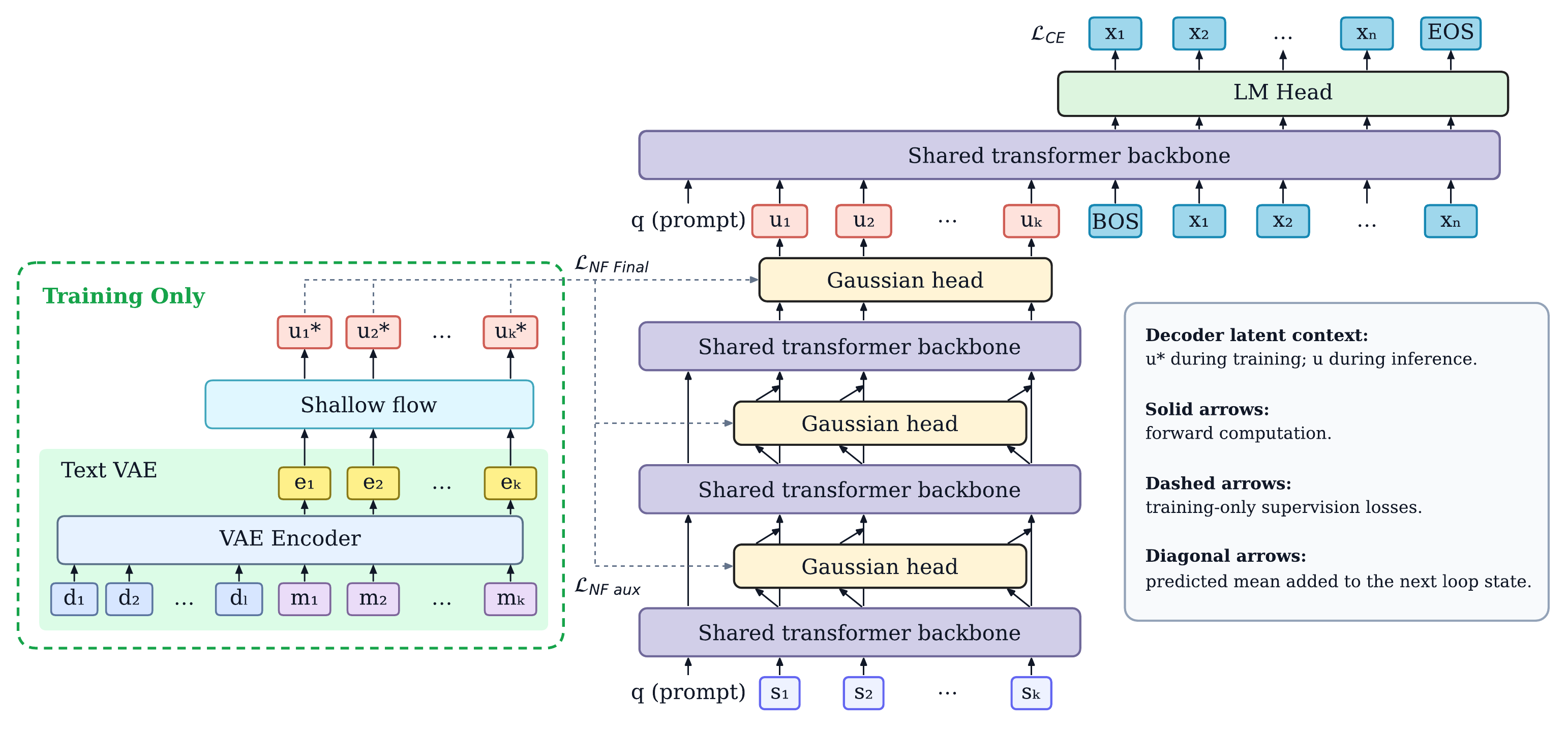}%
  }
  \caption{\textbf{Overview of the LLoCoT architecture.} A shared transformer repeatedly refines K loop hidden states, with each non-final Gaussian mean providing feedback to the next loop. The final Gaussian head generates all final latent thoughts jointly for autoregressive answer decoding; VAE-derived targets and auxiliary intermediate supervision are used only during training.}
  \label{fig:llocot_overview}
\end{figure}

\input{sections/related_works}

\input{sections/method}
\input{sections/results}
\input{sections/limitations_and_conclusion}

\subsection*{AI use statement}

Generative AI tools were used to assist with drafting and editing this
manuscript and with writing code. All reported experimental values were
recomputed from logged outputs and checked by an author, and all LLM-assisted
code was reviewed and tested before use. We take responsibility for the final
content of this work.

\subsection*{Ethics statement}

This work does not use personal or sensitive data. Unlike explicit
chain-of-thought systems, latent reasoning does not expose its intermediate
thoughts, which may reduce interpretability and make the model's reasoning
harder to inspect or monitor. Deployment should therefore include external
evaluation, testing, and appropriate safety controls.

\subsection*{Reproducibility statement}

We describe the model architecture, training objectives, loop and feedback
configurations, benchmark versions, prompts, decoding settings, evaluation
metrics, ablations, and inference-efficiency protocol in the paper and its
appendices. Accuracy is evaluated with EvalPlus on HumanEval, HumanEval+,
MBPP, and MBPP+, using fixed benchmark task sets and the reported evaluation
procedure. The efficiency measurements specify the prompts, hardware,
precision, batch size, warm-up procedure, and repetition protocol.

An accompanying repository is being prepared for open-source release. Once
cleaned up, it will include the implementation, training and evaluation
scripts, configuration files, dependency information, experiment metadata,
and documentation needed to reproduce the reported results. Where permitted
by the relevant licenses, we will also provide checkpoints or links to the
corresponding pretrained and derived models.


\bibliography{references}
\bibliographystyle{iclr2027_conference}

\clearpage

\appendix
\input{sections/appendix}

\end{document}

%% file: math_commands.tex
\usepackage{amsmath,amsfonts,bm}

\def\eqref#1{equation~\ref{#1}}

\def\1{\bm{1}}

\DeclareMathAlphabet{\mathsfit}{\encodingdefault}{\sfdefault}{m}{sl}
\SetMathAlphabet{\mathsfit}{bold}{\encodingdefault}{\sfdefault}{bx}{n}



%% file: sections/introduction.tex
\section{Introduction}

Test-time computation has become an important means of improving the
reasoning capabilities of language models. Rather than producing an answer
immediately, a model can devote additional computation to decomposing a
problem, developing intermediate conclusions, and revising its solution.
Chain-of-thought (CoT) reasoning realizes this process by generating an explicit
sequence of reasoning tokens before the final answer.
However, CoT ties reasoning computation to autoregressive language
generation, as every intermediate thought must be expressed in natural language
and generated after all preceding tokens. Increasing the amount of reasoning
therefore requires extending a strictly sequential trace, even when the
intermediate computation is needed only internally and does not need to be
communicated to the user.

Continuous latent reasoning relaxes the requirement that intermediate
computation be represented as text. A reasoning trace can be encoded into a
compact collection of continuous vectors that act as learned aggregators of
information from the complete trace. These vectors need not correspond to
individual words, sentences, or human-interpretable steps. However, most
existing latent-reasoning approaches generate their latent states
autoregressively. \textbf{LLoCoT}, our architecture, is motivated by the
intuition that latent thoughts summarizing a complete reasoning trace need not
follow a fixed linguistic order. Instead of forming a sequence that is extended
from left to right, they can constitute a shared workspace that is repeatedly
refined through parameter-shared computation, in the spirit of
TRM~\citep{trm}.

We instantiate this idea using continuous latent targets derived from
NF-CoT~\citep{nfcot}. LLoCoT initializes a fixed set of learned hidden-state
slots and repeatedly refines them with a shared transformer. At each non-final
iteration, the predicted latent means are projected back into the workspace to
guide the next refinement step. After the final iteration, all latent thoughts
are sampled jointly from the predicted Gaussian distribution and used to
condition autoregressive answer generation. The model is trained with a
likelihood objective on the final latent predictions, auxiliary supervision
that aligns intermediate predictions with the same complete-trace target, and
a cross-entropy objective for answer generation. A frozen VAE and a conditional
flow construct these continuous training targets, but neither is required at
inference time.

Across several coding benchmarks, we find that LLoCoT performs on par with
reasoning SFT while outperforming the base model, answer-only supervised
fine-tuning, and NF-CoT on aggregate. In particular, LLoCoT consistently
improves over NF-CoT, indicating that iterative refinement of a shared latent
space provides an effective alternative to autoregressive latent-thought
generation. In terms of efficiency, LLoCoT substantially shortens the reasoning
phase relative to both textual chain-of-thought and autoregressive latent
reasoning. Its parallelizable refinement process brings the time to the first
answer token close to that of direct answer generation and yields the lowest
end-to-end latency among the evaluated methods.

We summarize our contributions as follows:
\begin{itemize}\itemsep2pt
\item We introduce LLoCoT, an architecture that replaces autoregressive
latent-thought generation with recurrent refinement of a fixed-size latent
workspace.
\item We show that LLoCoT performs on par with Reasoning SFT while
outperforming the base model, answer-only supervised fine-tuning and NF-CoT.
\item We show that LLoCoT substantially reduces reasoning latency relative to
both textual and autoregressive latent reasoning, yielding the lowest
end-to-end latency among the evaluated methods.
\end{itemize}

%% file: sections/related_works.tex
\section{Related works}

\textbf{Latent reasoning.}
Chain-of-thought prompting improves language-model reasoning by allocating additional computation through intermediate textual tokens \citep{cot}. Latent-reasoning methods move this computation into continuous hidden states, avoiding the requirement that every intermediate step be verbalized. Coconut \citep{coconut} introduced an early instance of this paradigm by feeding a model's hidden state directly back as its next input embedding. Subsequent approaches have explored alternative generative models for continuous thoughts. LaDiR \citep{ladir}, for example, encodes reasoning traces with a variational autoencoder (VAE) and iteratively denoises blocks of latent thoughts using diffusion, replacing left-to-right generation with non-autoregressive refinement. 
Most closely related to our work is NF-CoT \citep{nfcot}, which encodes explicit reasoning traces into compact VAE representations and uses normalizing flows to construct a tractable distribution over continuous thoughts. NF-CoT preserves probabilistic sampling and likelihood-based training, but generates its latent thoughts autoregressively. We build on its VAE- and flow-based target construction while replacing sequential latent generation with the joint, iterative refinement of a fixed latent workspace.

\textbf{Looped transformers.}
Looped transformers repeatedly apply a shared block to an evolving hidden state, decoupling computational depth from parameter count. Prior empirical work has shown that this recurrence can support iterative state manipulation and the execution of algorithmic procedures \citep{saunshi2025reasoninglatentthoughtspower,kohli2026loopthinkgeneralize}. More recently, HRM \citep{hrm} and TRM \citep{trm} demonstrated that recurrent refinement can also solve highly structured reasoning problems such as Sudoku, maze planning, and ARC-AGI with relatively small models. 
Although these results are obtained in constrained domains, they suggest that repeatedly updating a latent state provides a useful inductive bias for multi-step computation. Along a complementary line of work, looped architectures have also begun to scale to general-purpose language modeling \citep{ouro, nanbeige}.

%% file: sections/method.tex
\section{LLoCoT: Latent Looped Chain-of-Thought}
\label{sec:method}

Let $q$ denote an input prompt, $d$ an explicit reasoning trace available
during training, and $y$ the corresponding final answer. At inference, the
model receives only $q$. LLoCoT replaces left-to-right latent-thought
generation with iterative refinement of a fixed set of learned workspace
slots. The same transformer is reapplied for $R$ refinement iterations, and
the resulting latent thoughts condition an autoregressive answer decoder. The
overall computation is illustrated in Figure~\ref{fig:llocot_overview}.

The model maintains $K$ hidden-state slots of width $d_h$, where $d_h$ is the
hidden dimension of the transformer backbone. It predicts $K$ final latent
thoughts of width $D$ and uses them to condition answer generation. The main
configuration uses $K=64$, $d_h=2048$, $D=2560$, and $R=2$; additional
hyperparameters are provided in Appendix~\ref{app:implementation_details}.

\subsection{NF-CoT-Derived Targets}
\label{sec:method_targets}

Following NF-CoT~\citep{nfcot}, we construct continuous training targets from
the explicit reasoning trace using a separately trained transformer-based VAE.
The VAE encoder receives the reasoning trace $d$ and maps it to a sequence of
$K$ latent vectors,
\[
e^\star_{1:K}
=
\operatorname{stopgrad}
\left(
E_{\mathrm{VAE}}(d)
\right)
\in
\mathbb{R}^{K\times D},
\]
where the stored representation is the approximate posterior mean. The $K$
vectors jointly summarize the complete reasoning trace; they are not assigned
to individual tokens, sentences, or fixed reasoning steps. The VAE encoder is
pretrained separately and remains frozen throughout LLoCoT training. The
prompt $q$ and final answer $y$ are not inputs to the VAE encoder. Here,
$\operatorname{stopgrad}$ indicates that no gradients are propagated into the
VAE parameters.

During LLoCoT training, isotropic dequantization noise is added to the VAE
representation,
\[
\widetilde{e}^\star_{1:K}
=
e^\star_{1:K}+\eta,
\qquad
\eta\sim\mathcal{N}(0,\tau^2 I),
\]
with $\tau=0.3$ in the reported configuration. A conditional shallow
autoregressive flow then maps the perturbed VAE representation into the
LLM-facing final-latent-thought space,
\[
u^\star_{1:K}
=
F_{\phi}\!\left(
\widetilde{e}^\star_{1:K};q
\right)
\in
\mathbb{R}^{K\times D}.
\]
The flow parameters $\phi$ are trained as part of the target-space adapter
during LLoCoT training. Its initialization and stage-wise optimization
details are provided in Appendix~\ref{app:implementation_details}.

The transformed representation $u^\star_{1:K}$ serves as the continuous
latent target for LLoCoT training and as the conditioning sequence for
target-conditioned answer training. The target-space flow is an adapter
between the VAE coordinate system and the LLM-facing latent coordinate system;
it should not be confused with NF-CoT's autoregressive latent generator. Both
the VAE and the target-space flow are used only during training and are not
executed during inference.

\subsection{Looped Hidden-State Refinement and Mean Feedback}
\label{sec:method_loop_refinement}

The loop hidden states are initialized from $K$ prompt-independent,
trainable slot vectors,
\[
H^0=S\in\mathbb{R}^{K\times d_h}.
\]
These vectors are initialized once, optimized during training, and reused as
the initial workspace for every prompt. They are not derived from the
reasoning trace.

At each refinement iteration, the current hidden states are combined with the
prompt representation and processed by the shared transformer backbone,
\[
\widetilde{H}^{r}
=
\mathcal{T}_{\theta}(H^{r-1};q),
\qquad
r=1,\ldots,R,
\]
where $\mathcal{T}_{\theta}$ denotes the shared transformer computation. The
same transformer parameters are reused across iterations, allowing the same
latent slots to be progressively refined without introducing a separate
transformer for each loop. Each transformer invocation updates all $K$
hidden-state slots in parallel under the selected attention mask; the only
loop-level recurrence is between refinement iterations.

A shared slotwise Gaussian head is applied to every refined hidden-state
sequence,
\[
(\mu^r_{1:K},\sigma^r_{1:K})
=
G_{\theta}(\widetilde{H}^{r}),
\qquad
\mu^r_{1:K},\sigma^r_{1:K}
\in
\mathbb{R}^{K\times D},
\qquad
r=1,\ldots,R,
\]
where the predicted scales are constrained to be positive. For every
non-final iteration, the predicted mean is mapped into the transformer hidden
dimension and added to the refined hidden states,
\[
H^r
=
\widetilde{H}^{r}
+
P_{\theta}\!\left(\mu^r_{1:K}\right),
\qquad
r=1,\ldots,R-1.
\]
The final Gaussian parameters are defined by
\[
(\mu_{1:K},\sigma_{1:K})
\equiv
(\mu^R_{1:K},\sigma^R_{1:K}),
\]
since no feedback update is applied after the final iteration.

The projection $P_{\theta}$ is shared across slots and iterations and is
applied independently to each slot. It maps the $D$-dimensional
final-latent-thought space to the $d_h$-dimensional hidden-state space and is
also used to condition the answer decoder. Mean feedback changes the
refinement computation itself: each intermediate prediction is made
available to the next loop, allowing later iterations to refine hidden states
that already contain an estimate of the final latent thoughts. Using the
predicted mean rather than a sampled latent avoids injecting additional
sampling noise into this refinement trajectory. The default loop-attention
mask is causal over the hidden-state slots.

All $K$ slots are updated in parallel within each refinement iteration, whereas
the refinement iterations remain sequential because each iteration depends on
the previous workspace. Hyperparameter values are provided in
Appendix~\ref{app:implementation_details}.

\subsection{Final Latent Thoughts, Training Objective, and Inference}
\label{sec:method_objective}

We use $\theta$ for the shared transformer, seed-slot, Gaussian-head, and
latent-projection parameters; $\phi$ for the target-space flow parameters;
and $\psi$ for any answer-decoder-specific parameters. Shared transformer
parameters used by the answer decoder are included in $\theta$. The VAE
parameters are denoted by $\omega$ and remain fixed.

The final Gaussian head defines a factorized distribution over the final
latent thoughts,
\[
p_{\theta}(u_{1:K}\mid q)
=
\prod_{i=1}^{K}
\mathcal{N}
\left(
u_i;
\mu_i(q),
\operatorname{diag}\!\left(\sigma_i(q)^2\right)
\right).
\]
Although the distribution factorizes across slots and latent dimensions, its
parameters are produced from jointly refined hidden states and therefore
depend on $q$ through the refinement process.

Let $M=KD$, and let $J_{F_{\phi},e}$ denote the Jacobian of the target-space
flow with respect to its latent input, holding the prompt fixed. Because
\[
u^\star_{1:K}
=
F_{\phi}\!\left(
\widetilde{e}^\star_{1:K};q
\right),
\]
the change-of-variables formula gives the induced density in the perturbed
VAE coordinates:
\[
p_{\theta,\phi}^{(e)}
\!\left(
\widetilde{e}^\star_{1:K}\mid q
\right)
=
p_{\theta}\!\left(u^\star_{1:K}\mid q\right)
\left|
\det
J_{F_{\phi},e}
\!\left(
\widetilde{e}^\star_{1:K};q
\right)
\right|.
\]
Because the target-space flow is autoregressive, its Jacobian is triangular
and the log-determinant is computed from the blockwise scale terms rather than
from a dense $KD\times KD$ determinant. This computation is required only
during training.

We therefore use the following per-element negative log-likelihood:
\[
\mathcal{L}_{\mathrm{final}}(\theta,\phi)
=
-\frac{1}{M}
\log p_{\theta}\!\left(u^\star_{1:K}\mid q\right)
-
\frac{1}{M}
\log
\left|
\det
J_{F_{\phi},e}
\!\left(
\widetilde{e}^\star_{1:K};q
\right)
\right|.
\]
The constant Gaussian normalization term is omitted. Since the VAE
representation is detached, no gradients are propagated into the VAE
parameters. Gradients from the primary latent objective may update the
LLoCoT parameters and the trainable target-space flow.

For $R>1$, the non-final Gaussian predictions are trained against the same
target final latent thoughts,
\[
\mathcal{L}_{\mathrm{intermediate}}(\theta,\phi)
=
\frac{1}{R-1}
\sum_{r=1}^{R-1}
\left[
-\frac{1}{M}
\log
\mathcal{N}
\left(
\operatorname{stopgrad}(u^\star_{1:K});
\mu^r_{1:K},
\operatorname{diag}
\left((\sigma^r_{1:K})^2\right)
\right)
\right],
\]
where the transformed target is detached in this auxiliary term. For $R=1$,
we define $\mathcal{L}_{\mathrm{intermediate}}=0$. Intermediate supervision
aligns the non-final predictions with the continuous target, whereas mean
feedback allows those predictions to influence subsequent refinement. This
auxiliary term is evaluated directly in the transformed final-latent-thought
coordinates; the flow Jacobian is therefore required only for the primary
latent likelihood.

For an answer containing $T$ tokens, target-conditioned training uses
$u^{\mathrm{ctx}}_{1:K}=u^\star_{1:K}$ and the autoregressive answer loss
\[
\mathcal{L}_{\mathrm{answer}}(\theta,\phi,\psi)
=
-\frac{1}{T}
\sum_{t=1}^{T}
\log
p_{\theta,\psi}
\left(
y_t
\mid
y_{<t},q,u^\star_{1:K}
\right).
\]
The target latent sequence is treated as a conditioning target; no gradients
are propagated into the VAE encoder. The complete training objective is
\[
\mathcal{L}(\theta,\phi,\psi;\omega)
=
\lambda_{\mathrm{NF}}
\left(
\mathcal{L}_{\mathrm{final}}(\theta,\phi)
+
\alpha_{\mathrm{intermediate}}
\mathcal{L}_{\mathrm{intermediate}}(\theta,\phi)
\right)
+
\lambda_{\mathrm{CE}}
\mathcal{L}_{\mathrm{answer}}(\theta,\phi,\psi),
\]
where $\omega$ is fixed and the remaining parameters are optimized according
to the stage-specific training procedure described in
Appendix~\ref{app:implementation_details}.

At inference time, the model receives only the prompt $q$. It initializes the
loop hidden states, applies the shared transformer for $R$ refinement
iterations, and applies mean feedback after each non-final iteration. The
final Gaussian parameters are then used to obtain all $K$ final latent
thoughts jointly:
\[
u_{1:K}
=
\mu_{1:K}
+
\sigma_{1:K}\odot\epsilon,
\qquad
\epsilon\sim\mathcal{N}(0,I),
\]
for stochastic decoding, or
\[
u_{1:K}=\mu_{1:K}
\]
for deterministic decoding.

The selected latent thoughts are projected into the decoder hidden dimension
and used to generate the answer autoregressively. Answer decoding begins only
after the final latent thoughts have been produced and projected. Target
construction, intermediate supervision, the VAE, the target-space flow, and
answer teacher forcing are absent during inference.

%% file: sections/results.tex
\section{Results}
\label{sec:results}

We evaluate LLoCoT against supervised and latent-reasoning baselines, followed by ablations of its refinement mechanism and an analysis of its computational efficiency.
We evaluate HumanEval, HumanEval+, MBPP, and MBPP+ with EvalPlus~\citep{evalplus}.
All benchmark entries report percentages with 95\% fixed-task sampling
intervals. Mean columns are descriptive arithmetic averages of the four
reported benchmark point estimates and do not carry an additional confidence
interval. Detailed decoding protocols and evaluator validation are provided in Appendix~\ref{app:results_protocol}.

\subsection{Accuracy Results}
\label{sec:results_main}

Table~\ref{tab:results_main} compares LLoCoT with the base model, supervised
fine-tuning controls, and NF-CoT. We report results separately on the original
and EvalPlus versions of HumanEval and MBPP, together with their unweighted
mean.

LLoCoT achieves the highest four-benchmark mean of 63.6. This represents an
average improvement of 6.3 points over the Qwen3-1.7B base model, 0.9 points
over Base SFT, and 1.6 points over NF-CoT. LLoCoT obtains higher point
estimates than NF-CoT on all four benchmarks, with the largest gains appearing
on MBPP and MBPP+ at 2.9 and 3.3 percentage points, respectively. The final
LLoCoT configuration, which combines mean feedback with auxiliary intermediate
supervision, raises the average from 63.1 for the no-feedback configuration to
63.6, with the largest individual gain appearing on MBPP+.

Reasoning SFT is the strongest competing method by average performance,
achieving a mean of 63.2. It performs best on HumanEval and HumanEval+,
whereas LLoCoT performs best on MBPP and MBPP+. Overall, LLoCoT performs on
par with Reasoning SFT and outperforms the base model, Base SFT, NF-CoT, and
the no-feedback LLoCoT configuration.

\begin{table}[H]
\centering
\scriptsize
\setlength{\tabcolsep}{3pt}
\caption{Main code-generation results. Values are percentages. Benchmark
entries are point estimate $\pm$ the 95\% fixed-task sampling-interval
half-width. The mean is the unweighted arithmetic mean of the four benchmark
point estimates. Parenthetical values in the Mean column are changes relative
to Qwen3-1.7B-Base. Best and second-best available entries are shown in
\textbf{bold} and \underline{underlined}, respectively.}
\label{tab:results_main}
\resizebox{0.9\textwidth}{!}{%
\begin{tabular}{@{}lccccc@{}}
\toprule
Mode
& HumanEval
& HumanEval+
& MBPP
& MBPP+
& Mean \\
\midrule
Qwen3-1.7B-Base
& $59.9 \pm 1.2$
& $52.6 \pm 1.3$
& $63.3 \pm 0.8$
& $53.4 \pm 0.8$
& $57.3$ \resneutral{+0.0} \\

Base SFT
& $68.6 \pm 1.1$
& $\underline{63.9 \pm 1.1}$
& $63.4 \pm 0.7$
& $\underline{55.0 \pm 0.6}$
& $62.7$ \resgood{+5.4} \\

Reasoning SFT
& $\mathbf{70.3 \pm 1.0}$
& $\mathbf{64.4 \pm 1.0}$
& $63.5 \pm 0.7$
& $54.7 \pm 0.7$
& $\underline{63.2}$ \resgood{+5.9} \\

\midrule

NF-CoT
& $69.2 \pm 1.1$
& $63.5 \pm 1.1$
& $62.2 \pm 0.7$
& $52.9 \pm 0.7$
& $62.0$ \resgood{+4.7} \\

No-feedback LLoCoT, $R=2$
& $\underline{69.3 \pm 1.1}$
& $63.8 \pm 1.1$
& $\underline{64.3 \pm 0.7}$
& $\underline{55.0 \pm 0.6}$
& $63.1$ \resgood{+5.8} \\

LLoCoT
& $\underline{69.3 \pm 1.0}$
& $\underline{63.9 \pm 1.0}$
& $\mathbf{65.1 \pm 0.6}$
& $\mathbf{56.2 \pm 0.6}$
& $\mathbf{63.6}$ \resgood{+6.3} \\
\bottomrule
\end{tabular}%
}
\end{table}

\subsection{Selected Ablations}
\label{sec:results_ablations}

We focus the main-text ablations on two questions: how performance changes
with the number of refinement loops, and which feedback or conditioning
mechanism provides the strongest results.

\paragraph{Loop depth.}

Table~\ref{tab:results_loop_depth_main} varies the number of causal
no-feedback refinement loops. Performance is not monotonic in loop depth:
the strongest four-benchmark mean occurs at a relatively small number of
loops, while deeper configurations do not consistently improve either
benchmark family.

\begin{table}[H]
\centering
\scriptsize
\setlength{\tabcolsep}{3pt}
\caption{Loop-depth ablation for No-feedback LLoCoT. Values are percentages
and report point estimate $\pm$ the 95\% fixed-task sampling-interval
half-width. The mean is the arithmetic average of the four benchmark point
estimates. Parenthetical values in the mean column are changes relative to
No-feedback LLoCoT, $R=2$. Green, red, and gray indicate positive, negative,
and zero changes, respectively.}
\label{tab:results_loop_depth_main}
\resizebox{0.9\textwidth}{!}{%
\begin{tabular}{@{}lccccc@{}}
\toprule
Mode
& HumanEval
& HumanEval+
& MBPP
& MBPP+
& Mean \\
\midrule
No-feedback LLoCoT, $R=1$
& $69.1 \pm 1.0$
& $63.3 \pm 1.1$
& $\mathbf{64.4 \pm 0.7}$
& $\underline{54.9 \pm 0.6}$
& $62.9$ \resbad{-0.2} \\

No-feedback LLoCoT, $R=2$
& $\underline{69.3 \pm 1.1}$
& $\underline{63.8 \pm 1.1}$
& $\underline{64.3 \pm 0.7}$
& $\mathbf{55.0 \pm 0.6}$
& $\underline{63.1}$ \resneutral{+0.0} \\

No-feedback LLoCoT, $R=3$
& $\mathbf{69.4 \pm 1.0}$
& $\mathbf{64.0 \pm 1.1}$
& $\underline{64.3 \pm 0.6}$
& $\mathbf{55.0 \pm 0.6}$
& $\mathbf{63.2}$ \resgood{+0.1} \\

No-feedback LLoCoT, $R=4$
& $68.2 \pm 1.0$
& $63.1 \pm 1.0$
& $63.8 \pm 0.7$
& $54.8 \pm 0.6$
& $62.5$ \resbad{-0.6} \\

No-feedback LLoCoT, $R=6$
& $68.6 \pm 1.1$
& $63.6 \pm 1.1$
& $60.4 \pm 0.7$
& $52.0 \pm 0.6$
& $61.2$ \resbad{-1.9} \\

No-feedback LLoCoT, $R=8$
& $68.0 \pm 1.0$
& $63.0 \pm 1.0$
& $62.2 \pm 0.6$
& $53.6 \pm 0.6$
& $61.7$ \resbad{-1.4} \\
\bottomrule
\end{tabular}
}
\end{table}

\paragraph{Feedback and conditioning.}

We compare feedback aggregation, answer conditioning, and auxiliary
supervision while keeping the causal workspace and two-loop configuration
fixed. Each variant is constructed from the No-feedback LLoCoT control.

\begin{itemize}
    \item \textbf{No-feedback LLoCoT, $R=2$.}
    This is the matched control without feedback aggregation. It serves as the
    reference point for the remaining comparisons.

    \item \textbf{Answer-mean.}
    This variant conditions the answer decoder on the model's predicted latent
    mean rather than the teacher-forced target latent. It tests whether answer
    generation remains effective when the decoder receives the model's own
    latent prediction.

    \item \textbf{Auxiliary-NLL loss.}
    This variant adds an auxiliary negative-log-likelihood objective during
    training. It tests whether additional supervision at intermediate
    refinement steps improves the learned latent representation.

    \item \textbf{Feedback-mean + auxiliary NLL (LLoCoT).}
    This is the final LLoCoT configuration, combining mean feedback with
    auxiliary supervision of the non-final Gaussian prediction in the
    two-loop causal workspace.

    \item \textbf{Feedback-sampled + auxiliary NLL.}
    This variant uses the same auxiliary supervision as the final
    configuration but samples the feedback representation instead of using its
    mean. It isolates the effect of stochastic rather than deterministic
    feedback.
\end{itemize}

Table~\ref{tab:results_feedback_main} reports the corresponding results. The
results show that how intermediate latent predictions are used is important.
Conditioning the answer decoder directly on the predicted latent mean causes a
large performance drop, suggesting a substantial mismatch between predicted
and target latent representations. Auxiliary intermediate supervision alone
shifts performance toward the MBPP benchmarks but does not improve the overall
mean. Combining this supervision with mean feedback yields the strongest
result, improving the no-feedback baseline while preserving performance across
both benchmark families. Replacing mean feedback with sampled feedback
reduces the overall mean.

\begin{table}[H]
\centering
\scriptsize
\setlength{\tabcolsep}{3pt}
\caption{Feedback, conditioning, and auxiliary-supervision ablations. Values
are percentages and report point estimate $\pm$ the 95\% fixed-task sampling
interval half-width. The mean is the arithmetic average of the four benchmark
point estimates. Parenthetical values in the mean column are changes relative
to No-feedback LLoCoT, $R=2$. Green, red, and gray indicate positive, negative,
and zero changes, respectively.}
\label{tab:results_feedback_main}
\resizebox{0.9\textwidth}{!}{%
\begin{tabular}{@{}lccccc@{}}
\toprule
Mode
& HumanEval
& HumanEval+
& MBPP
& MBPP+
& Mean \\
\midrule
No-feedback LLoCoT, $R=2$
& $\mathbf{69.3 \pm 1.1}$
& $\underline{63.8 \pm 1.1}$
& $64.3 \pm 0.7$
& $55.0 \pm 0.6$
& $63.1$ \resneutral{+0.0} \\

+ Answer-mean
& $61.7 \pm 1.2$
& $55.3 \pm 1.2$
& $60.2 \pm 0.7$
& $51.6 \pm 0.7$
& $57.2$ \resbad{-5.9} \\


+ Auxiliary-NLL loss
& $68.1 \pm 1.1$
& $62.6 \pm 1.0$
& $65.0 \pm 0.7$
& $56.0 \pm 0.6$
& $62.9$ \resbad{-0.2} \\

+ Feedback-mean + auxiliary NLL (LLoCoT)
& $\mathbf{69.3 \pm 1.0}$
& $\mathbf{63.9 \pm 1.0}$
& $\underline{65.1 \pm 0.6}$
& $\underline{56.2 \pm 0.6}$
& $\underline{63.6}$ \resgood{+0.5} \\

+ Feedback-sampled + auxiliary NLL
& $67.9 \pm 1.1$
& $62.3 \pm 1.0$
& $64.2 \pm 0.6$
& $55.3 \pm 0.6$
& $62.4$ \resbad{-0.7} \\


\bottomrule
\end{tabular}
}
\end{table}

\subsection{Inference Efficiency}
\label{sec:results_efficiency}

Table~\ref{tab:results_efficiency} reports matched sampled-efficiency
measurements for the base model, Reasoning SFT, NF-CoT, and LLoCoT.
Measurements use 128 fixed coding prompts, batch size one, a single NVIDIA
H100 80GB HBM3 GPU, and bfloat16. Each prompt is measured three times after
ten warm-up runs, yielding 384 measured records per endpoint. The efficiency
answer-length limit is 2048 tokens and is separate from the accuracy-generation
cap. Warm-up, tokenization, sampling, network, and benchmark-evaluation time
are excluded.
For Reasoning SFT, the reasoning phase ends at the first
\texttt{</think>} boundary. For NF-CoT and LLoCoT, the corresponding phase is
serial latent generation or looped latent refinement. Conditioning or latent
selection and answer decoding are measured separately. Peak allocated memory
and total transformer forward calls are reported as representative memory and
invocation measures. Analytical FLOPs are deterministic estimates.

\begin{table}[H]
\centering
\scriptsize
\setlength{\tabcolsep}{4pt}
\caption{Sampled inference-efficiency measurements. Values are point
estimates. Parenthetical multiplicative ratios for NF-CoT and LLoCoT are
relative to Reasoning SFT. Ratios are rounded to the displayed precision;
values below $0.1$ retain one significant digit. Best and second-best values
in each row are shown in bold and underlined, respectively. Colors indicate
favorable, unfavorable, and neutral changes relative to Reasoning SFT.}
\label{tab:results_efficiency}
\resizebox{0.85\textwidth}{!}{%
\begin{tabular}{@{}lcccc@{}}
\toprule
Metric
& Base model
& Reason. SFT
& NF-CoT
& LLoCoT \\
\midrule
Time to first answer token (s) $\downarrow$
& $\mathbf{0.03}$
& $3.17$
& $2.01$ \resgood{$\times 0.6$}
& $\underline{0.09}$ \resgood{$\times 0.03$} \\

End-to-end latency (s) $\downarrow$
& $22.59$
& $15.16$
& $\underline{13.73}$ \resgood{$\times 0.9$}
& $\mathbf{11.47}$ \resgood{$\times 0.8$} \\

Serial reasoning/latent phase (s) $\downarrow$
& $\mathbf{0}$
& $2.47$
& $1.98$ \resgood{$\times 0.8$}
& $\underline{0.06}$ \resgood{$\times 0.02$} \\

Conditioning/selection phase (s) $\downarrow$
& $\mathbf{0}$
& $0.70$
& $\underline{0.03}$ \resgood{$\times 0.04$}
& $\underline{0.03}$ \resgood{$\times 0.04$} \\

Peak allocated (GiB) $\downarrow$
& $\underline{3.4}$
& $\mathbf{3.3}$
& $6.3$ \resbad{$\times 1.9$}
& $6.3$ \resbad{$\times 1.9$} \\

Reasoning/latent forward calls $\downarrow$
& $\mathbf{1.0}$
& $81.5$
& $64.0$ \resgood{$\times 0.8$}
& $\underline{2.0}$ \resgood{$\times 0.02$} \\

Total transformer forward calls $\downarrow$
& $768.6$
& $514.4$
& $\underline{457.1}$ \resgood{$\times 0.9$}
& $\mathbf{403.1}$ \resgood{$\times 0.8$} \\

Analytical FLOPs ($\times 10^{12}$) $\downarrow$
& $\underline{2.6}$
& $\mathbf{1.7}$
& $\mathbf{1.7}$ \resneutral{$\times 1.0$}
& $2.8$ \resbad{$\times 1.6$} \\
\bottomrule
\end{tabular}
}
\end{table}

The table clearly shows that LLoCoT substantially reduces the latency of the reasoning phase. Its time to first answer token is only 0.09 seconds, compared with 3.17 seconds for Reasoning SFT and 2.01 seconds for NF-CoT, making it the fastest reasoning method and bringing its startup latency close to that of the base model, which performs no explicit reasoning. Similarly, LLoCoT completes its latent refinement phase in 0.06 seconds, over $40\times$ faster than the textual reasoning phase of Reasoning SFT and over $30\times$ faster than the autoregressive latent generation of NF-CoT. This reduction translates into the lowest end-to-end latency. LLoCoT also requires the fewest transformer forward calls, although its peak memory matches NF-CoT at 6.3 GiB and its analytical FLOP count is higher than those of the other reasoning methods. Thus, LLoCoT's wall-clock advantage arises primarily from replacing serial latent generation with more parallelizable refinement, rather than from reducing total arithmetic computation.

%% file: sections/limitations_and_conclusion.tex
\section{Limitations and Conclusions}







\subsection{Limitations}
Our study has several limitations. First, all experiments use a 1.7B-parameter backbone, so it remains unclear whether the observed benefits extend to larger models. Second, the evaluation is limited to code-generation tasks and does not establish generalization to other reasoning domains. Third, LLoCoT uses more peak memory than Reasoning SFT and more analytical FLOPs than the other evaluated reasoning methods; its speedup therefore results from replacing serial reasoning with more parallelizable computation rather than reducing total computational cost. Finally, as with many latent-reasoning approaches, the workspace size and refinement depth must be selected in advance rather than adapted dynamically to the difficulty of each input.
  
\subsection{Conclusions}

We introduce LLoCoT, a latent-reasoning architecture that replaces autoregressive latent-thought generation with recurrent refinement of a fixed-size latent state. LLoCoT repeatedly applies a shared transformer, feeds intermediate latent predictions back into subsequent refinement iterations, and jointly generates the final latent thoughts used to condition answer decoding. This design combines complete-trace continuous targets with a more parallelizable reasoning process.
Across four code-generation benchmarks, LLoCoT performs on par with Reasoning SFT while outperforming the base model, answer-only supervised fine-tuning, and NF-CoT on aggregate. It also obtains higher point estimates than NF-CoT on every evaluated benchmark. Most notably, LLoCoT substantially shortens the reasoning phase relative to both textual chain-of-thought and autoregressive latent reasoning, resulting in the lowest measured end-to-end latency. These results suggest that recurrent latent-workspace refinement is a promising approach for preserving the benefits of learned latent reasoning while reducing its sequential inference bottleneck.

%% file: sections/appendix.tex
\appendix

\section{Implementation Details}
\label{app:implementation_details}

Unless otherwise stated, the values in this appendix correspond to the
reported $R=2$ LLoCoT model with mean feedback.

\subsection{Target VAE}

The continuous targets were produced with a transformer-based VAE
implementation of the NF-CoT target interface. Both the encoder and decoder
use \texttt{Qwen/Qwen3-1.7B-Base}, whose hidden width is $2048$. The encoder is
causal, uses $K=64$ learned query slots, and predicts a
$D=2560$-dimensional latent vector for each slot. The VAE is not
prompt-conditioned and does not use block segmentation.

The VAE objective uses $\beta=0.01$, with posterior log-variances clipped to
$[-10,10]$. VAE training uses AdamW with learning rate $10^{-5}$ and
weight decay $0.01$, a per-GPU batch size of $4$, gradient accumulation of
$4$, two GPUs, and an effective batch size of $32$. Training uses bfloat16
precision, gradient checkpointing, a maximum sequence length of $3119$, seed
$0$, and $180{,}030$ optimizer steps.

Latent extraction is deterministic and stores posterior means. These cached
posterior means constitute the fixed VAE-derived target representation. The
VAE encoder remains frozen during all subsequent LLoCoT training. The VAE
decoder is not used during LLoCoT training or inference. During LLoCoT
training, Gaussian dequantization noise with standard deviation $0.3$ is added
to the cached representation before target-space flow transformation.

\subsection{LLoCoT Architecture}

The answer model is \texttt{Qwen/Qwen3-1.7B-Base}, with hidden width
$d_h=2048$, $28$ transformer layers, $16$ attention heads, and $8$ key--value
heads. LLoCoT uses $K=64$ loop hidden-state slots, final latent thoughts of
width $D=2560$, and $R=2$ refinement loops. The full transformer backbone is
reapplied at every loop using causal attention. The learned initial loop
hidden states are initialized from $\mathcal{N}(0,0.02^2)$, and stochastic
seed noise is disabled.

The latent head is a shared diagonal Gaussian head applied independently to
each slot. It maps each hidden-state vector to $2D$ outputs, corresponding to
the mean and scale of one final latent thought. The head is zero-initialized.
Means and log-scales use a soft-clipped transformation with clipping value
$4.0$, and scales are computed with softplus followed by a minimum value of
$10^{-5}$.

Mean feedback is applied after every non-final loop. The intermediate
predicted mean is projected into the hidden dimension with
\[
\operatorname{LayerNorm}(D)
\rightarrow
\operatorname{Linear}(D,d_h)
\rightarrow
\operatorname{GELU}
\rightarrow
\operatorname{Linear}(d_h,d_h),
\]
and added to the next loop hidden states. This projection is shared with the
projection used to condition answer decoding. The intermediate Gaussian
objective has weight $0.25$ and is applied only to non-final loops, with the
target treated as stop-gradient. Prompt dropout is applied with probability
$0.1$.

The target-space flow contains five causal affine blocks, two conditioner
layers per block, a conditioner hidden size of $2048$, and $32$ attention
heads. The latent-slot ordering alternates between identity and reversal
across blocks. The flow is trained as a target-space adapter during LLoCoT
training and is not executed during inference.

\subsection{Parameter Status and Training Stages}

The VAE encoder is pretrained separately and remains frozen during all LLoCoT
stages. The target-space flow is not separately pretrained. Each affine block
is initialized to the identity transformation by zero-initializing its shift
and log-scale output projection. The flow is then optimized jointly with the
other trainable LLoCoT modules during both LLoCoT training stages.

Stage~1 freezes the Qwen transformer backbone. The loop-specific modules,
latent Gaussian head, target-space flow, latent projection, and learned
initial hidden states are trained. Stage~2 is initialized from the Stage~1
checkpoint and unfreezes the Qwen backbone. The target-space flow and
loop-specific modules remain trainable unless explicitly disabled by the
experiment configuration.

The VAE-derived representation is treated as fixed with respect to the VAE
parameters. Gradients through the target-space flow are retained when the
flow is trainable. The intermediate latent target is detached for the
auxiliary intermediate objective.

\subsection{Optimization}

Both LLoCoT stages use a per-GPU batch size of $16$, gradient accumulation of
$1$, four GPUs, and an effective global batch size of $64$. Training uses
bfloat16 precision, gradient checkpointing, gradient clipping at $1.0$, and
weight decay $0.01$. The optimizer is AdamW with
$\beta_1=0.9$ and $\beta_2=0.95$.

Stage~1 uses learning rate $10^{-4}$, $100$ warmup steps, seed $17$, and
$1563$ optimizer steps. Stage~2 is initialized from the Stage~1 checkpoint,
uses learning rate $5\times10^{-5}$, $200$ warmup steps, seed $0$, and
$45{,}222$ optimizer steps. Both stages use cosine learning-rate decay.

The loss weights are $\lambda_{\mathrm{NF}}=1$ and
$\lambda_{\mathrm{CE}}=1$. The intermediate latent objective uses weight
$\alpha_{\mathrm{intermediate}}=0.25$. The reported model uses target-latent
answer conditioning during training. Conditioning variants using predicted
means or sampled latents are evaluated separately.

The maximum prompt and answer lengths during training are both $1024$ tokens.
These training limits are distinct from the benchmark-specific generation
settings described in Appendix~\ref{app:results_protocol}.

\subsection{Generation}

For reported stochastic accuracy evaluations, latent sampling uses
temperature $0.9$ and CFG scale $1.0$. Answer generation uses temperature
$0.6$ and top-$p=1.0$. All final latent thoughts are generated jointly; only
the answer decoder remains autoregressive.

Accuracy evaluations use $n=16$ completions per task with distinct stable
seeds. EvalPlus answer generation uses a maximum of $2048$ answer tokens.
Native-thinking models use a maximum reasoning length of $8192$ tokens and the
same $2048$-token answer cap.

\section{Evaluation Protocol and Audit}
\label{app:results_protocol}

\subsection{Benchmark and Decoding Protocol}

We use the HumanEval and MBPP task sets through EvalPlus~\citep{evalplus} and
report results on HumanEval, HumanEval+, MBPP, and MBPP+. All models use the
same fixed task sets, answer-extraction procedure, and benchmark evaluator.
Prompt rendering follows the model-specific protocol described below.

Each stochastic evaluation generates $n=16$ completions per task using
distinct stable seeds. Ordinary code-SFT and generated-latent models use the
EvalPlus stock-wrapped prompt,
\texttt{evalplus-stock-wrapped}, implemented with
\texttt{make\_raw\_chat\_prompt}. Native-thinking models use the corresponding
two-phase reasoning-and-answer protocol.

Answer generation uses temperature $0.6$, top-$p=1.0$, and a maximum of
$2048$ answer tokens. Native-thinking models use a maximum reasoning length of
$8192$ tokens. Generated-latent NF-CoT and LLoCoT use latent temperature
$0.9$ and CFG scale $1.0$.

\subsection{Failure Handling and Integrity Checks}

Every expected task--sample pair produces an evaluation record. Invalid,
empty, truncated, malformed, evaluator-rejected, and execution-failed
completions are retained as score-zero outcomes and remain in the fixed
$n=16$ denominator.

Missing or duplicate task--sample records, invalid task identities, corrupted
evaluator artifacts, and inconsistent evaluator outputs are treated as
integrity failures and block result promotion. They are not silently repaired,
discarded, or replaced.

\subsection{Pass@1 and Confidence Intervals}

For each task $t$, let $c_t$ denote the number of successful completions among
the $n=16$ samples. The reported pass@1 score is the average of the per-task
success rates,
\[
\operatorname{pass@1}
=
\frac{1}{|\mathcal{T}|}
\sum_{t\in\mathcal{T}}
\frac{c_t}{16}.
\]

Each stochastic accuracy entry is reported as a point estimate with a
symmetric approximate 95\% fixed-task sampling interval. The interval
half-width is computed using the fixed-task plug-in Wald procedure. These
intervals quantify sampling uncertainty over the fixed benchmark tasks, not
uncertainty over task selection, training seeds, or model selection.
Deterministic evaluations receive point estimates without sampling intervals.

Paired intervals are used only when task identities, sample identities, and
seeds are aligned. The mean reported in the main results table is the
unweighted arithmetic mean of the four benchmark point estimates:
\[
\operatorname{Mean}
=
\frac{
s_{\mathrm{HumanEval}}
+
s_{\mathrm{HumanEval+}}
+
s_{\mathrm{MBPP}}
+
s_{\mathrm{MBPP+}}
}{4}.
\]
This descriptive mean does not receive an additional confidence interval.

\subsection{Evaluation Records and Reproducibility}

Each evaluation records the benchmark version, task identity, sample identity,
stable seed, prompt variant, decoding parameters, model checkpoint, evaluator
output, and completion status. Results are promoted only after all expected
records are present, sample identities are unique, evaluator artifacts are
valid, and the reported point estimates and interval artifacts are complete.

\subsection{Inference-Efficiency Protocol}

Efficiency measurements use 128 fixed coding prompts, batch size one, a single
NVIDIA H100 80GB HBM3 GPU, and bfloat16. Each prompt is measured three times
after ten warm-up runs, yielding $384$ measured records per endpoint. The
efficiency answer-length limit is $2048$ tokens and is separate from the
accuracy-generation cap. Warm-up, tokenization, sampling, network, and
benchmark-evaluation time are excluded.

For Reasoning SFT, the reasoning phase ends at the first
\texttt{</think>} boundary. For NF-CoT and LLoCoT, the corresponding phase is
serial latent generation or looped latent refinement. Conditioning or latent
selection and answer decoding are measured separately when applicable.

Time to first answer token is measured from the start of model execution to
the first answer token. End-to-end latency includes the measured reasoning,
conditioning, and answer-generation phases. Peak allocated memory is recorded
during each measured run. Analytical FLOPs are deterministic estimates.

Forward-pass counts are recorded directly by the evaluation counter and are
averaged over the measured prompt--run records; consequently, they may be
non-integer. Reasoning or latent forward calls and total transformer forward
calls are reported separately. These counts describe invocation structure and
do not replace analytical FLOP estimates.